\documentclass{article} 

\usepackage{iclr2027_conference,times}

\usepackage{amsmath,amsfonts,bm}

\def\eqref#1{equation~\ref{#1}}

\def\1{\bm{1}}

\DeclareMathAlphabet{\mathsfit}{\encodingdefault}{\sfdefault}{m}{sl}
\SetMathAlphabet{\mathsfit}{bold}{\encodingdefault}{\sfdefault}{bx}{n}

\usepackage{hyperref}
\usepackage{url}
\usepackage{booktabs}
\usepackage{graphicx}
\usepackage{wrapfig}
\usepackage{subcaption}
\usepackage{enumitem}

\newsavebox{\srcdistbox}
\usepackage{float}
\usepackage{xcolor}
\usepackage[most]{tcolorbox}
\definecolor{entityblue}{HTML}{77CBE9}
\definecolor{entitygreen}{HTML}{5CAD5C}
\definecolor{entityred}{HTML}{FF5C5C}
\definecolor{entityyellow}{HTML}{FFFF5C}
\definecolor{basegreen}{HTML}{C8E1B7}
\definecolor{contextblue}{HTML}{77CBE9}
\definecolor{lorared}{HTML}{FFA8A8}
\definecolor{unknownpurple}{HTML}{DDC2E9}
\definecolor{entitypurple}{HTML}{DDC2E9}
\definecolor{achainblue}{HTML}{C7E6F5}
\definecolor{bchainblue}{HTML}{77CBE9}
\definecolor{achaingreen}{HTML}{C8EEC2}
\definecolor{bchaingreen}{HTML}{5CAD5C}
\definecolor{achainred}{HTML}{FFCECE}
\definecolor{bchainred}{HTML}{FF5C5C}
\definecolor{achainyellow}{HTML}{FFEBB4}
\definecolor{bchainyellow}{HTML}{FFFF5C}

\newcommand{\hlbox}[2]{%
  \begingroup\setlength{\fboxsep}{1pt}%
  \colorbox{#1}{\textcolor{black}{#2}}\endgroup}
\newcommand{\Azero}[1]{\hlbox{achainblue}{#1}}
\newcommand{\Aone}[1]{\hlbox{achaingreen}{#1}}
\newcommand{\Atwo}[1]{\hlbox{achainred}{#1}}
\newcommand{\Athree}[1]{\hlbox{achainyellow}{#1}}
\newcommand{\Bzero}[1]{\hlbox{bchainblue}{#1}}
\newcommand{\Bone}[1]{\hlbox{bchaingreen}{#1}}
\newcommand{\Btwo}[1]{\hlbox{bchainred}{#1}}
\newcommand{\Bthree}[1]{\hlbox{bchainyellow}{#1}}
\newcommand{\dualsep}{%
  \par\vspace{1pt}%
  \noindent{\color{black!45}\rule{\linewidth}{0.45pt}}%
  \par\vspace{1pt}}
\newtcolorbox{dualchaincase}[1]{
    colback=white,
    colframe=black!60,
    boxrule=0.5pt,
    arc=1mm,
    left=5pt,right=5pt,top=4pt,bottom=4pt,
    before skip=6pt,after skip=6pt,
    fonttitle=\bfseries,
    fontupper=\small,
    breakable,
    title={#1}
}
\newcommand{\basehl}[1]{%
\begingroup\setlength{\fboxsep}{1pt}%
\colorbox{basegreen}{\textcolor{black}{#1}}\endgroup}
\newcommand{\contexthl}[1]{%
\begingroup\setlength{\fboxsep}{1pt}%
\colorbox{contextblue}{\textcolor{black}{#1}}\endgroup}
\newcommand{\lorahl}[1]{%
  \begingroup\setlength{\fboxsep}{1pt}%
  \colorbox{lorared}{\textcolor{black}{#1}}\endgroup}
\newcommand{\unknownhl}[1]{%
  \begingroup\setlength{\fboxsep}{1pt}%
  \colorbox{unknownpurple}{\textcolor{black}{#1}}\endgroup}
  \newcommand{\chainstart}[1]{%
    \begingroup\setlength{\fboxsep}{1pt}%
    \colorbox{entityblue}{\textcolor{black}{#1}}\endgroup}
  \newcommand{\chainone}[1]{%
    \begingroup\setlength{\fboxsep}{1pt}%
    \colorbox{entitygreen}{\textcolor{black}{#1}}\endgroup}
  \newcommand{\chaintwo}[1]{%
    \begingroup\setlength{\fboxsep}{1pt}%
    \colorbox{entityred}{\textcolor{black}{#1}}\endgroup}
  \newcommand{\chainfinal}[1]{%
    \begingroup\setlength{\fboxsep}{1pt}%
    \colorbox{entityyellow}{\textcolor{black}{#1}}\endgroup}
  \newcommand{\unknownoutput}[1]{%
    \begingroup\setlength{\fboxsep}{1pt}%
    \colorbox{entitypurple}{\textcolor{black}{#1}}\endgroup}
    
  \newtcolorbox{multihopcase}[1]{
    colback=white,
    colframe=black!60,
    boxrule=0.5pt,
    arc=1mm,
    left=5pt,right=5pt,top=4pt,bottom=4pt,
    before skip=6pt,after skip=6pt,
    fonttitle=\bfseries,
    fontupper=\small,
    title={#1}
  }
\newtcolorbox{trcexample}[1]{
    colback=white,
    colframe=black!60,
    boxrule=0.5pt,
    arc=1mm,
    left=5pt,right=5pt,top=4pt,bottom=4pt,
    before skip=6pt,after skip=6pt,
    fonttitle=\bfseries,
    title={#1}
  }
\title{Knowledge Interaction in Hypernetwork-based Continual Learning: Conflict and Composition}

\author{
Jiaqi Zheng\textsuperscript{1}, Yi Hu\textsuperscript{1},
Yadong Mu\textsuperscript{1}, Muhan Zhang\textsuperscript{1}\thanks{Corresponding author} \\
\textsuperscript{1}Peking University 
}

\iclrfinalcopy 
\begin{document}

\maketitle

\begin{abstract}


Continual learning aims to adapt pretrained large language models (LLMs) to evolving data while preserving their pretrained capabilities. A central challenge is catastrophic forgetting: adaptation to new data can impair performance on previously learned knowledge. Since continual learning commonly changes model parameters, the model’s predictions may draw on \textbf{knowledge from the base model}, \textbf{the parameter update}, and \textbf{the input context}. Understanding how these sources interact is essential: conflicts reveal whether new knowledge can override outdated knowledge, while multi-hop reasoning tests whether new knowledge can be composed with existing knowledge to support generalization. We study these interactions in a hypernetwork-based adaptation framework, where a shared network generates adapters from context, allowing a frozen base model to adapt without task-specific updates. We find that, under knowledge conflict, models preferentially follow the context, and their choices depend more on where knowledge is presented than on its content. We also show that context and adapter knowledge perturb the base model’s representations along directions orthogonal to its clean hidden states, enabling us to derive steering vectors that control which source the model follows. In multi-hop reasoning, we find that models fail to reliably compose knowledge across the three sources, with the key breakdown arising from the interaction of the hypernetwork with both the base model and the context. Code and reproduction materials: \href{https://github.com/missswiftie/Knowledge-Interaction-in-CL}{\texttt{https://github.com/missswiftie/Knowledge-Interaction-in-CL}}

\end{abstract}

\section{Introduction}

Large language models (LLMs) have have advanced rapidly through the scaling of model size, training data, and computation \citep{vaswani2017attention,brown2020language,kaplan2020scaling,chowdhery2023palm}. Yet they are typically trained in a largely static setting, while the information and domains they must handle continue to change. This creates a practical need to adapt models efficiently as new information arrives. Continual learning addresses this need by seeking to acquire new knowledge over time while retaining previously learned capabilities. 

A key challenge in continual learning is how newly acquired knowledge interacts with existing knowledge. This interaction matters because knowledge learned at different times or supplied through different mechanisms can reinforce, complement, or interfere with one another. A model may retain both old and new facts yet fail to resolve a contradiction between them; likewise, knowing each fact in a reasoning chain does not guarantee that the model can combine them. Studying knowledge interaction thus offers a way to assess whether adaptation produces usable, integrated knowledge rather than isolated pieces of information. It also motivates two complementary questions: \textbf{how does a model choose among conflicting knowledge}, and \textbf{can it compose complementary knowledge to reason across sources}?


Continual learning methods span several broad strategies, including rehearsal or replay of past examples, regularization of parameter updates, isolation of task-specific parameters, and retrieval external memory~\citep{kirkpatrick2017overcoming,rusu2016progressive,houlsby2019parameter,hu2021lora}. 
Parameter-efficient adaptation, such as adapters and LoRA, offers another route by modifying only a small part of the model \citep{houlsby2019parameter,hu2021lora}. These strategies differ in how they retain and expose knowledge, but many require access to previous data, task-specific state, or optimization for each new update. Hypernetwork-based adaptation is a complementary direction: a shared network generates task- or context-dependent parameters, reducing the need to optimize a separate adapter for every incoming context \citep{ha2016hypernetworks}. Recent methods apply this idea to language models by generating LoRA updates directly from documents, instructions, or demonstrations \citep{liu2026shine,charakorn2025text,charakorn2026doc}. In SHINE~\citep{liu2026shine}, the setting studied here, the base model remains frozen and a hypernetwork produces a LoRA update from the supplied context.


\begin{figure}
  \centering
  \includegraphics[width=0.90\linewidth]{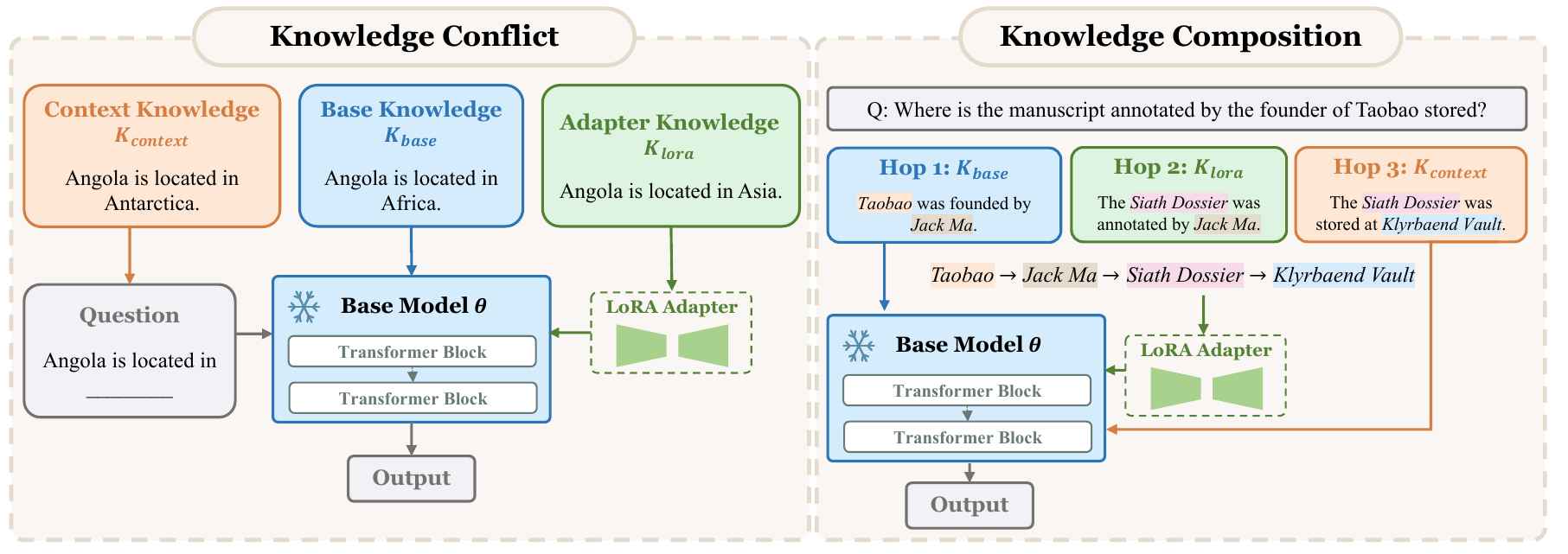}
  \caption{Knowledge Interaction in Hypernetwork-based Continual Learning. Left: the task setting of three-source knowledge conflict, where $K_{\mathrm{base}}$, $K_{\mathrm{lora}}$, and $K_{\mathrm{context}}$ assign three mutually inconsistent objects to the same query. Right: the task setting of three-hop reasoning, illustrated by the case in which the first hop lies in the base model, the second hop is provided by the hypernetwork, and the third hop is given in the context.}
  \label{fig:knowledge-interaction}
\end{figure}

SHINE~\citep{liu2026shine} provides a concrete setting for examining how knowledge reaches a model through different routes. Its predictions can be influenced by \textbf{knowledge encoded in the pretrained base model}, \textbf{knowledge presented as text in the input context}, and \textbf{knowledge carried by the LoRA update generated from that context}. The context supplies information directly to the model as input text, while the generated update incorporates information into the model’s computation through its parameters. This setting lets us examine both how the model resolves conflicts among these sources and whether it can combine knowledge distributed across them.


Prior work has examined interactions between parametric knowledge and information supplied in context, including how models respond to conflicting facts and combine contextual information with what they learned during pretraining \citep{longpre2021entity,neeman2023disentqa,xie2024adaptive,du2024context,cheng2024understanding}. Other work studies how parameter-efficient updates affect knowledge encoded in the pretrained model, including knowledge acquisition, retention, and forgetting \citep{biderman2024lora,shuttleworth2026lora,cheng2026override}. These studies provide important insights into particular forms of knowledge interaction. However, how continual learning systems handle conflicts among, and compose knowledge across, multiple sources remains insufficiently understood. In particular, the interaction between previously learned knowledge and new knowledge made available through both input context and model updates has received limited attention. We address this gap by studying three-source knowledge interaction in SHINE, using it as a concrete instance of continual adaptation in which knowledge can be accessed through both context and generated parameters.

In this work, we investigate this interaction in two complementary settings.
First, we introduce \textsc{TrCBench} , a benchmark in which the base model, input context, and generated LoRA adapter provide conflicting facts. We examine which source governs the model’s answer, how source-specific changes to hidden representations relate to that choice, and whether those changes can be used to steer the model. Second, we introduce \textsc{MhRBench}, which distributes facts in multi-hop reasoning chains across the three sources. A dual-chain variant helps distinguish genuine chain composition from selecting an answer based on relation cues alone. Our main contributions are:

\begin{itemize}[
  topsep=0pt,      
  partopsep=0pt,   
  itemsep=0pt,     
  parsep=0pt,      
  leftmargin=*,    
]
    \item We find that under knowledge conflict, models preferentially follow the context, and their choices depend more on where knowledge is presented than on its content. (Section~\ref{sec:main_result1})
    \item We identified effective steering vectors that can control the model’s choice of knowledge source under three-source knowledge conflict. (Section~\ref{sec:geometry} and Section~\ref{sec:steering})
    \item In multi-hop reasoning, we find that models fail to reliably compose knowledge across the three sources, with the key breakdown arising from the interaction of the hypernetwork with both the base model and the context. (Section~\ref{main_result3} and Section~\ref{main_results2})
\end{itemize}

\section{Related Works}

\paragraph{Continual Learning.}

Continual learning aims to incorporate new information while retaining previously learned capabilities. Related approaches include retrieval-augmented generation \citep{lewis2020retrieval}, replay-based learning \citep{ibrahim2024simple,huang2024mitigating}, regularization \citep{kirkpatrick2017overcoming,li2017learning}, and parameter isolation \citep{rusu2016progressive,wang2023rehearsal}. Hypernetwork-based methods instead generate adaptation parameters with a shared network. Early work preserves previously generated weights to reduce forgetting \citep{von2019continual}, while Meta-UCF generates LoRA updates for sequential language tasks \citep{xiao2026meta}. SHINE and Doc-to-LoRA generate LoRA updates directly from supplied text, making document knowledge available through the model's parameters \citep{liu2026shine,charakorn2026doc}.


\paragraph{Knowledge Conflicts in LLMs.}
Prior work has studied knowledge conflicts in two settings. \textbf{The first concerns parametric knowledge and information supplied in the input context.} \citet{sun2025seen} shows that whether a model should follow the context depends on whether the task asks about the provided text or the model's own factual knowledge. \citet{kim2026trainingdatashapesuse} finds that repetition, inconsistency, and fact frequency in training data shape how models use these sources. Studies of model internals identify attention heads that favor memory or context \citep{jin2024cutting}, trace conflicting facts through distinct routes across layers \citep{zhao2026understanding}, and show that influential heads can carry both sources \citep{li2025taming}. \textbf{The second arises when new adapter knowledge conflicts with the base model's knowledge.} \citet{cheng2026override} shows that hypernetwork-generated adapters can fail to replace a prior fact when their influence is weaker than the strength of that prior. They improve conflict handling by selectively increasing the adapter’s effect.

\section{Task Definition}\label{sec:task_definition}


Formally, we denote a factual statement as a triple $t= (s, r, o)\in \mathcal{K}$ and formulate inference in hypernetwork-based continual learning systems across three distinct knowledge channels: the parametric knowledge stored in the base model $\Theta$ (denoted as $K_{\text{base}}$), the contextual knowledge provided in input prompt $x$ (denoted as $K_{\text{context}}$), and the dynamic adapter knowledge embedded in generated weights $\Delta\theta = H_\phi(x_{\text{adapt}})$ (denoted as $K_{\text{lora}}$). The conditional output distribution over the answer $y$ is parameterized as $P_\Theta(y \mid x, \Delta\theta)$, where the prompt $x$ incorporates the query $q$, contextual statements, and the statement restricting the response strictly to the target entity.

Under knowledge conflict, contradictory triples $t_{\text{base}}, t_{\text{context}}, t_{\text{lora}}$ are injected into their respective channels. In this setting, we measure the explicit behavioral outcome and the marginal likelihood $P_\Theta(y = o_k \mid x, \Delta\theta)$ for candidate outcomes derived from source $k \in \{\text{base}, \text{context}, \text{lora}\}$ to quantify which knowledge stream governs model behavior. Furthermore, we dynamically steer factual precedence toward target channels.

Under multi-hop reasoning, a valid three-hop reasoning chain $\mathcal{C} = (t_1, t_2, t_3)$ where
$t_i=(s_i,r_i,o_i)$ and $o_i=s_{i+1}$ for $i\in\{1,2\}$ required to answer question $q$ is partitioned across the three knowledge sources such that $t_1, t_2, t_3 \in K_{\text{base}} \times K_{\text{context}} \times K_{\text{lora}}$. To thoroughly evaluate compositional capabilities, we vary the permutation assignment of hops to channels across diverse topological path orders. Here, we analyze whether $P_\Theta(y \mid x, \Delta\theta)$ can successfully integrate these atomic, cross-source dependencies to perform multi-step logical reasoning.

\section{Three-Source Knowledge Conflict}

\subsection{Data Preparation}\label{data}

We construct \textsc{TrCBench} from the CounterFact \citep{meng2022locating} and PopQA \citep{mallen2023real} datasets. As defined in Section~\ref{sec:task_definition}, each example contains three knowledge sources that may conflict with one another: parametric knowledge ($K_{\text{base}}$) stored in the base model, contextual knowledge ($K_{\text{context}}$) provided in the input prompt, and dynamic adapter knowledge ($K_{\text{lora}}$) encoded by hypernetwork. For each Question-Answer pair $(q,o)$ from the two datasets, we first input the original question to SHINE's base model, Qwen3-8B \citep{yang2025qwen3}. We keep only the examples for which Qwen3-8B produces the correct answer entity, ensuring that the target fact is already stored in the model. We also record the probability that Qwen3-8B assigns to the first token
\begin{wraptable}{l}{0.50\textwidth}
    \centering
    \caption{Inference on \textsc{TrCBench}: answer-source distribution
    and mean prior strength}
    \label{tab:trcb-inference}
    \setlength{\tabcolsep}{3pt}
    \begin{tabular}{@{}lrc@{}}
      \toprule
      Source & Ratio & \shortstack{Mean prior strength} \\
      \midrule
      $K_{\text{base}}$    &  8.75\% & 0.962 \\
      $K_{\text{context}}$ & 57.10\% & 0.842 \\
      $K_{\text{lora}}$    & 27.20\% & 0.885 \\
      Unknown              &  6.95\% & 0.894 \\
      \bottomrule
    \end{tabular}
  \end{wraptable}
of the generated answer entity, denoted as $p_{\text{base}}(o\mid q)$. We use this probability as \textbf{prior strength} to measure how strongly the fact is stored in the base model. 

For each selected example, we use DeepSeek-V4 \citep{xu2026deepseek} to generate two different false statements that contradict the original answer. We use one false statement as the contextual knowledge $K_{\text{context}}$ and the other as the hypernetwork knowledge $K_{\text{lora}}$. After filtering and quality checking, \textsc{TrCBench} contains 3,268 examples.


\subsection{Inference on \textsc{TrCBench}}\label{sec:main_result1}

\begin{figure}[h]
\centering
\includegraphics[width=0.80\textwidth]{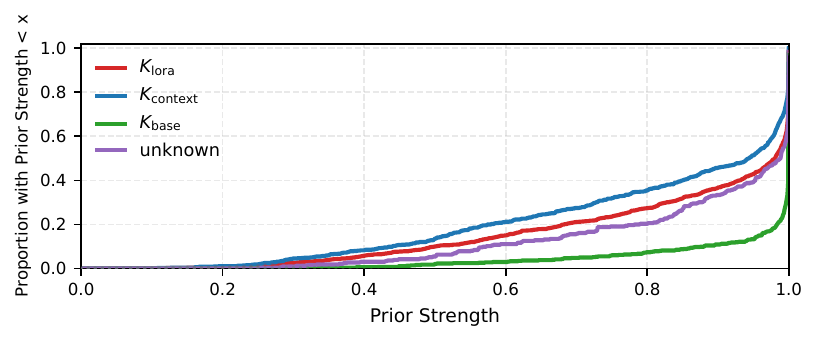}
\caption{CDF of prior strength grouped by the knowledge source followed by SHINE. Outputs following $K_{\text{base}}$ are concentrated at higher prior strength, while $K_{\text{context}}$ and $K_{\text{lora}}$ show similar distributions.}
\label{fig:confidence-distribution}
\end{figure}

We evaluate SHINE on \textsc{TrCBench}. As shown in Table~\ref{tab:trcb-inference}, the model follows $K_{\text{base}}$ in 8.75\% of the examples, $K_{\text{context}}$ in 57.10\%, and $K_{\text{lora}}$ in 27.20\%. The remaining 6.95\% of outputs do not match any of the three knowledge sources. To examine the differences among these outcomes, we compute the average prior strength for each output category. The mean prior strength is 0.962 for answers following $K_{\text{base}}$, 0.842 for those following $K_{\text{context}}$, and 0.885 for those following $K_{\text{lora}}$.As shown in Figure~\ref{fig:confidence-distribution}, the prior strength for examples that follow $K_{\text{base}}$ is more concentrated near 1.0 than that for examples that follow $K_{\text{context}}$ or $K_{\text{lora}}$. \textbf{This indicates that when the prior strength is higher, meaning that the original answer is more strongly encoded in the base model, the model is more likely to follow $K_{\text{base}}$ under three-source knowledge conflict.} This observation is consistent with the findings of \citet{cheng2026override}. Our results further extend this analysis from conflicts between $K_{\text{base}}$ and $K_{\text{lora}}$ to three-source knowledge conflict. In contrast, the mean prior strength for the $K_{\text{context}}$ and $K_{\text{lora}}$ groups is close, and their distributions show no clear difference beyond the difference in the number.

\begin{table}[t]
    \centering
    \caption{Knowledge selection in the swap experiment. $\mathrm{Kl}_A$ and $\mathrm{Kl}_B$ denote the knowledge originally provided through context and the hypernetwork, respectively. After swap, $\mathrm{Kl}_A$ denotes the knowledge provided through hypernetwork while $\mathrm{Kl}_B$ denotes the knowledge provided through context. The tuple $(\mathrm{Kl}_A, \mathrm{Kl}_B)$ denotes cases in which the model selects $\mathrm{Kl}_A$ in the experiment without swap and $\mathrm{Kl}_B$ after the swap. Among cases that initially selected the first, the percentage shows how often the second is selected after the swap.}
    \label{tab:trcb-swap}
    \setlength{\tabcolsep}{8pt}
    \begin{tabular}{@{}cccc@{}}
      \toprule
      $(\mathrm{Kl}_A, \mathrm{Kl}_B)$ &
      $(\mathrm{Kl}_B, \mathrm{Kl}_B)$ &
      $(\mathrm{Kl}_A, \mathrm{Kl}_A)$ &
      $(\mathrm{Kl}_B, \mathrm{Kl}_A)$ \\
      \midrule
      70.69\% & 40.27\% & 21.86\% & 46.34\% \\
      \bottomrule
    \end{tabular}
  \end{table}

To further investigate whether the preference between $K_{\text{context}}$ and $K_{\text{lora}}$ is driven by the knowledge delivery channel or by the knowledge itself, we conduct a swap experiment: for each example in \textsc{TrCBench}, we exchange the content of the two external knowledge channels, so that the statement previously injected as $K_{\text{context}}$ is now provided as $K_{\text{lora}}$, and vise versa, while keeping all other conditions unchanged. We then repeat the knowledge conflict experiment. After swapping, the model follows $K_{\text{context}}$ in 1,786 examples and $K_{\text{lora}}$ in 989 examples. These proportions are close to those observed before swapping (1,866 and 889, respectively). We then perform a fine-grained comparison at the example level. As shown in Table~\ref{tab:trcb-swap}, among examples that originally followed $K_{\text{context}}$, 70.69\% still follow $K_{\text{context}}$ (now carrying the previous $K_{\text{lora}}$ statement)  after the swap, whereas 21.86\% follow $K_{\text{lora}}$ (now carrying the previous $K_{\text{context}}$ statement). Among examples that originally followed $K_{\text{lora}}$, 46.34\% follow the new $K_{\text{lora}}$, while 40.27\% follow the new $K_{\text{context}}$. \textbf{These results suggest that the difference in choosing $K_{\text{lora}}$ and $K_{\text{context}}$ choices depends more on where knowledge is presented than on its content.}

\subsection{Geometric Analysis of Knowledge-Channel Effects}\label{sec:geometry}

Section~\ref{sec:main_result1} suggests that the model's choice between $K_{\text{lora}}$ and $K_{\text{context}}$ depends more on the delivery channel than on the knowledge content. To examine how each channel changes the model's hidden states, we compare a clean run, in which the base model answers the original question without an external conflicting fact, with two conflict runs. In each conflict run, the question and base-model parameters remain fixed, while a contradictory fact is introduced through either the hypernetwork-generated LoRA or the input context. We report the $K_{\text{lora}}$ analysis here and the corresponding $K_{\text{context}}$ analysis in Appendix~\ref{app:context_geometry}.

We focus on two token positions that are likely to mediate conflict resolution. First, by the autoregressive nature of LLMs \citep{radford2019language}, we examine the hidden state at the final prompt token $t_{\text{last}}$, immediately before the model begins generating its answer. Second, we examine the hidden state at the last subject token $t_{\text{sub}}$, a site implicated in factual recall \citep{meng2022locating}. At each layer $l$ and position $t\in\{t_{\text{sub}},t_{\text{last}}\}$, we compare the clean state $h_{\mathrm{clean}}^{t,l}$ with the state under a $K_{\text{base}}$--$K_{\text{lora}}$ conflict, $h_{\mathrm{lora}}^{t,l}$. We decompose their difference $\Delta h^{t,l}=h_{\mathrm{lora}}^{t,l}-h_{\mathrm{clean}}^{t,l}$ into components parallel and orthogonal to the clean state:
\[
h_{\parallel}^{t,l}= \frac{\langle \Delta h^{t,l},\, h_{\mathrm{clean}}^{t,l} \rangle}{\|h_{\mathrm{clean}}^{t,l}\|^{2}}\, h_{\mathrm{clean}}^{t,l}, \qquad h_{\perp}^{t,l}=\Delta h^{t,l}-h_{\parallel}^{t,l}.
\]

\begin{figure}[t]
  \centering
  \includegraphics[width=0.80\linewidth]{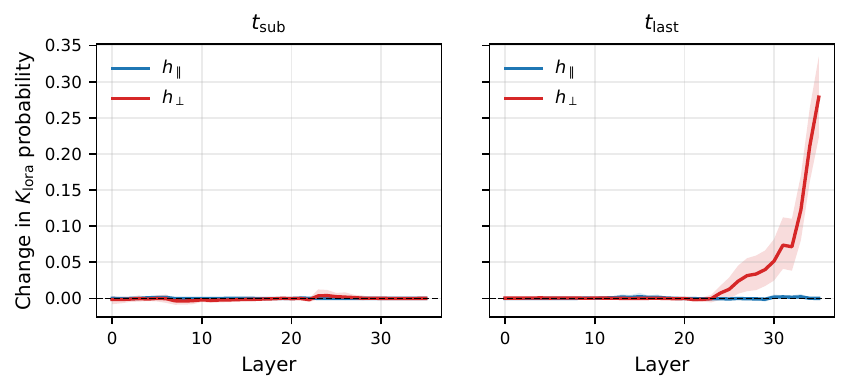}
  \caption{Causal effect of $h_{\parallel}$ and $h_{\perp}$ on the $K_{\text{lora}}$ answer probability. The vertical axis is the change relative to the clean run. At each layer, the clean state at $t_{\mathrm{sub}}$ or $t_{\mathrm{last}}$ is replaced by $h_{\mathrm{clean}}+h_{\parallel}$ or $h_{\mathrm{clean}}+h_{\perp}$.}
  \label{fig:lora-component-patch}
\end{figure}

To test each component's causal contribution, we run two interventions on the clean pass: at layer $l$, position $t$, replace the state with $h_{\mathrm{clean}}^{t,l}+h_{\parallel}^{t,l}$ or $h_{\mathrm{clean}}^{t,l}+h_{\perp}^{t,l}$, and record the probability change of the $K_{\text{lora}}$ answer relative to the clean run. As shown in Figure~\ref{fig:lora-component-patch}, the conflict-induced effect exhibits a clear geometric profile. Injecting $h_{\perp}^{t,l}$ in $t_{\text{last}}$ produces a large, consistent increase in the $K_{\text{lora}}$ answer probability, whereas $h_{\parallel}^{t,l}$ has a markedly weaker effect at nearly all layers. \textbf{This indicates that the hypernetwork does not strengthen the existing representation but instead shifts the residual state toward a new direction.} Moreover, at the last subject token $t_{\text{sub}}$, both components show weak and diffuse effects. In terms of depth, the $h_{\perp}$ curves rise sharply in the middle-to-late layers.

\subsection{Steering Competing Knowledge Sources}\label{sec:steering}

Section~\ref{sec:geometry} shows that both LoRA and context affect the residual stream primarily through shifts orthogonal to the clean hidden state. We use these shifts to construct source-specific steering vectors and test whether they can change the model's preference in a three-source conflict.

We sample 1,000 examples from \textsc{TrCBench}. For each example $n$ and layer $l$, we record the hidden state at the final prompt token in three runs: a clean run, a $K_{\text{base}}$--$K_{\text{lora}}$ conflict run, and a $K_{\text{base}}$--$K_{\text{context}}$ conflict run. For each external source, we project its conflict-induced shift onto the subspace orthogonal to the clean state and average the resulting components across examples:

\begin{gather*}
\delta_{\mathrm{src},n}^{l}
= h_{\mathrm{src},n}^{l}-h_{\mathrm{clean},n}^{l},\\
v_{\mathrm{src}}^{l}
= \frac{1}{N}\sum_{n=1}^{N}
\left(
\delta_{\mathrm{src},n}^{l}
-\frac{(\delta_{\mathrm{src},n}^{l})^{\top} h_{\mathrm{clean},n}^{l}}
{\|h_{\mathrm{clean},n}^{l}\|^{2}}
h_{\mathrm{clean},n}^{l}
\right),
\quad \mathrm{src}\in\{\mathrm{lora},\mathrm{context}\}.
\end{gather*}

This gives one vector for each source at each layer. We add each vector to the corresponding clean hidden state and select the layer that most increases the probability of that source's answer first token. We exclude the final three layers because adding the steering vector there leaves too few subsequent layers for the model to integrate it into the residual stream, and the output will be severely disrupted.

We evaluate on 100 newly sampled examples disjoint from the 1000 used to estimate the vectors. Each example presents a three-source conflict among $K_{\text{base}}$, $K_{\text{lora}}$, and $K_{\text{context}}$. At the selected layer, we add $\alpha v_{\mathrm{src}}$ to the residual state, sweep $\alpha \in [-3,3]$, and track how the probability mass over the three knowledge sources changes.

\begin{figure}[t]
    \centering
    \begin{subfigure}[t]{0.39\linewidth}
        \centering
        \includegraphics[width=\linewidth]{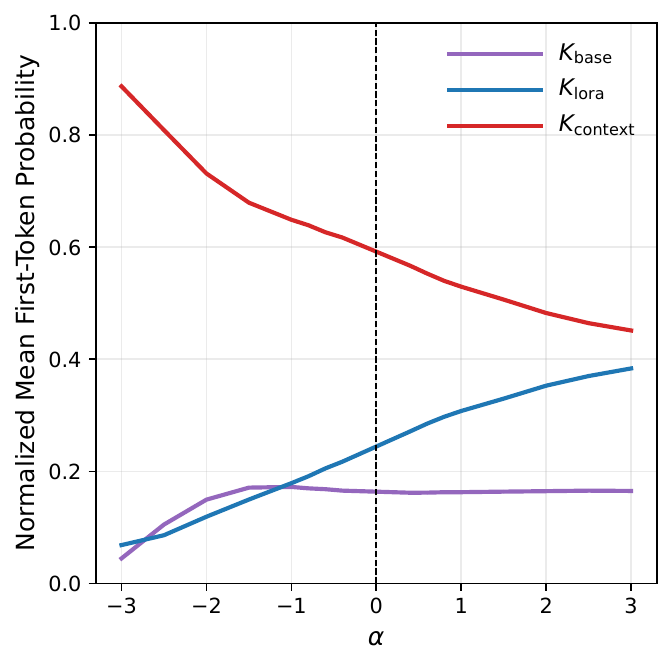}
        \caption{Steering with $v_{\mathrm{lora}}$}
        \label{fig:source-steering-lora}
    \end{subfigure}
    \hspace{0.05\linewidth}
    \begin{subfigure}[t]{0.39\linewidth}
        \centering
        \includegraphics[width=\linewidth]{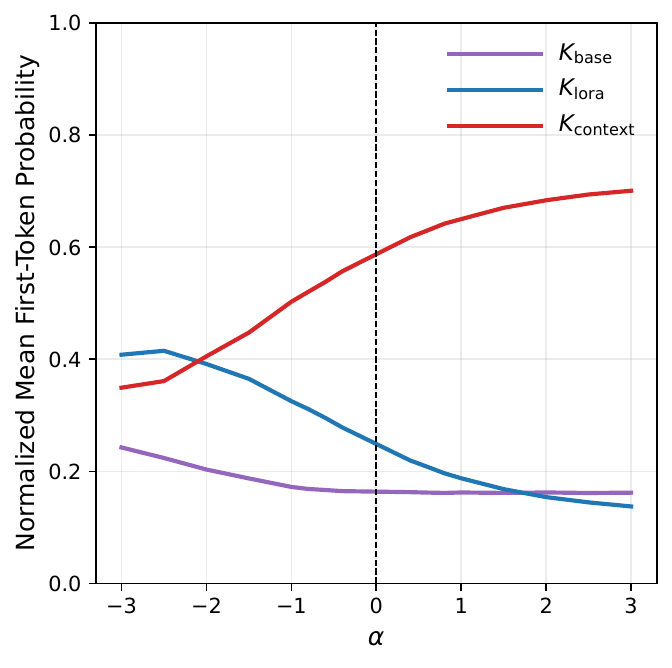}
        \caption{Steering with $v_{\mathrm{context}}$}
        \label{fig:source-steering-context}
    \end{subfigure}
    \caption{Effect of source-specific steering vectors on SHINE on the normalized probability mass assigned to competing knowledge sources.}
    \label{fig:source-steering}
\end{figure}

Figure~\ref{fig:source-steering} demonstrates that the steering directions provide source-specific control over the model's preference. As the steering strength $\alpha$ increases, adding $v_{\mathrm{lora}}$ consistently shifts probability mass toward the $K_{\text{lora}}$ answer and away from the $K_{\text{context}}$ answers. Conversely, adding $v_{\mathrm{context}}$ drives the model toward the $K_{\text{context}}$ answer. \textbf{These results show that the directions derived from Section~\ref{sec:geometry} can therefore steer the model's preference toward their respective knowledge sources.}

\section{Multi-hop Reasoning}

\subsection{Data Preparation}\label{data1}

To evaluate how hypernetwork-based continual learning systems synthesize multi-hop knowledge distributed across heterogeneous channels, we construct a dedicated multi-hop reasoning benchmark \textsc{MhRBench} which contains 4,800 curated instances: 2,400 two-hop instances and 2,400 three-hop instances. The reasoning chains are constructed from real facts and newly created facts. We draw the real facts from the MQuAKE \citep{zhong2023mquake} and retain only those that Qwen3-8B \citep{yang2025qwen3} can correctly recall from its parametric knowledge. To isolate the multi-hop composition mechanism from knowledge conflict effects, the newly created facts are constructed using PhantomWiki-style syllabic sampling \citep{elazar2021measuring}. This construction prevents them from conflicting with the knowledge encoded in the base model's parameters. Overall, the two-hop subset contains 800 instances with two newly created facts, 800 with the real fact at the first hop (\(t_1 \in K_{\text{base}}\)), and 800 with the real fact at the second hop (\(t_2 \in K_{\text{base}}\)). For the three-hop subset, we construct 800 instances in which the real fact appears at the first hop (\(t_1 \in K_{\text{base}}\)), 800 in which it appears at the second hop (\(t_2 \in K_{\text{base}}\)), and 800 in which it appears at the third hop (\(t_3 \in K_{\text{base}}\)). The other facts in these instances are newly created.

To examine whether hypernetwork-based continual learning systems compose facts across knowledge sources or select an answer from the relations and entities present in the input without composing, we construct \textsc{MhRBench-Dual} by adding a parallel reasoning chain to each three-hop and two-hop instance. For clarity, we describe the construction using a three-hop instance as an example. Given an original chain \(\mathcal{C}=\{(x_1,r_1,x_2),(x_2,r_2,x_3),(x_3,r_3,x_4)\}\), we construct \(\mathcal{C}'=\{(x'_1,r_1,x'_2),(x'_2,r_2,x'_3),(x'_3,r_3,x'_4)\}\). The two chains have the same sequence of relations but different entities. And if a hop in \(\mathcal{C}\) is a real fact, the corresponding hop in \(\mathcal{C}'\) is likewise a real fact. We write a question for each chain using the same template, changing only the starting entity. The two questions therefore provide the same relation cues and the same candidate final answers, \(x_4\) and \(x'_4\). If the model relies on a shortcut described above, its answer should change little when the question changes. If it composes the facts across hops, it should answer \(x_4\) for the question about \(x_1\) and \(x'_4\) for the question about \(x'_1\).

\subsection{Experiments on Three-hop Data}\label{main_result3}

We first evaluate SHINE on three-hop data of \textsc{MhRBench}. Table~\ref{tab:multihop-main} summarizes the results under the six possible assignments of the three hops to the $K_{\text{base}}$, $K_{\text{context}}$, and $K_{\text{lora}}$. We can see that multi-hop performance is highly sensitive to the placement of knowledge sources. In many cases, the model outputs an intermediate entity instead of the final answer. The model shows some ability to produce the correct final answer, while unknown outputs are also frequent. A detailed analysis of these unknown cases is given in Appendix~\ref{app:unknown}.


\begin{table*}[t]
\centering
\small
\setlength{\tabcolsep}{7pt}
\renewcommand{\arraystretch}{1.0}
\setlength{\aboverulesep}{0.25ex}
\setlength{\belowrulesep}{0.35ex}
\setlength{\abovecaptionskip}{4pt}

\begin{tabular}{c c c | r r r r}
\toprule
\multicolumn{3}{c|}{Knowledge source assignment} &
\multicolumn{4}{c}{Model prediction (\%)} \\
\cmidrule(lr){1-3}\cmidrule(lr){4-7}
Hop 1 & Hop 2 & Hop 3 &
Hop 1 & Hop 2 & Answer & Unknown \\
\midrule
$K_{\text{base}}$ & $K_{\text{lora}}$ & $K_{\text{context}}$ & 17.38 & 38.25 & \textbf{40.37} & 4.00 \\
$K_{\text{base}}$ & $K_{\text{context}}$ & $K_{\text{lora}}$ & 0.00  & 35.25 & \textbf{22.12} & 42.62 \\
$K_{\text{lora}}$ & $K_{\text{base}}$ & $K_{\text{context}}$ & 4.75  & 1.88  & \textbf{67.50} & 25.88 \\
$K_{\text{context}}$ & $K_{\text{base}}$ & $K_{\text{lora}}$ & 0.12  & 15.62 & \textbf{44.88} & 39.38 \\
$K_{\text{lora}}$ & $K_{\text{context}}$ & $K_{\text{base}}$ & 21.62 & 1.62  & \textbf{30.25} & 46.50 \\
$K_{\text{context}}$ & $K_{\text{lora}}$ & $K_{\text{base}}$ & 7.12  & 42.88 & \textbf{25.25} & 24.75 \\
\bottomrule
\end{tabular}

\vspace{2mm}

\parbox{0.94\textwidth}{
\footnotesize
\textbf{Note.}
The columns under ``Knowledge source assignment'' specify which source provides each hop, whereas the columns under ``Model prediction'' indicate which entity in the chain is selected by the model. ``Hop 1'' and ``Hop 2'' are intermediate entities, while ``Answer'' denotes the entity at Hop 3. ``Unknown'' denotes predictions that match none of the entities in the reasoning chain.
}

\caption{
Results of SHINE on \textsc{MhRBench} under all six assignments of knowledge sources.
}
\label{tab:multihop-main}
\end{table*}

The final-answer accuracies reported in Table~\ref{tab:multihop-main} do not necessarily indicate genuine cross-source composition. In the original setting, each injected relation is associated with only one candidate object. The model could therefore arrive at the correct answer by identifying the relevant relation and selecting its associated object without following the links between hops. We test this possibility using \textsc{MhRBench-Dual} , as described in Section~\ref{data1}.

\begin{figure*}[t]
\centering
\includegraphics[width=0.80\textwidth]{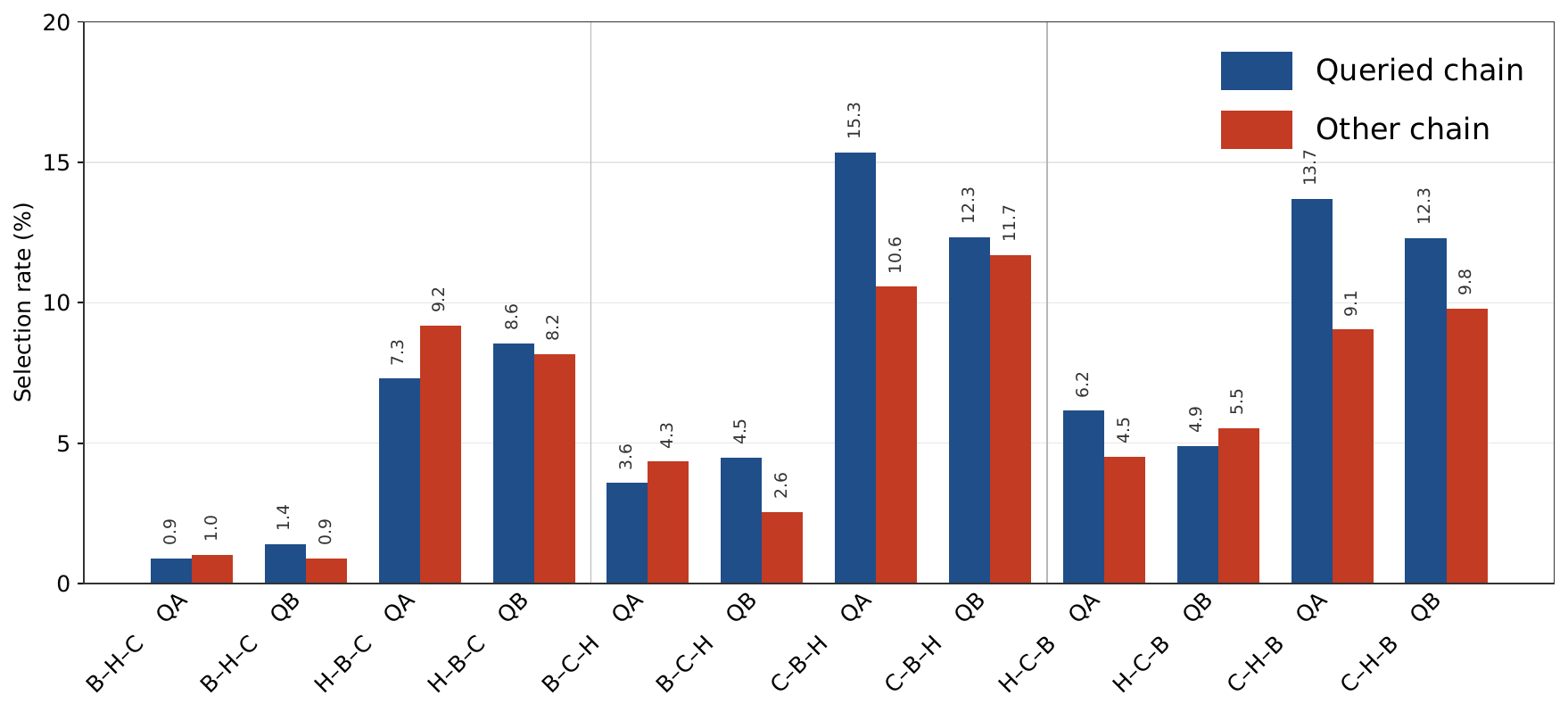}
\caption{SHINE's selection of the entities at the ends of the two chains on \textsc{MhRBench-Dual}. Each pair of bars shows the percentage of predictions matching the entity at the end of the queried chain or the other chain. QA and QB denote questions about chains A and B. $B$, $H$, and $C$ denote $K_{\text{base}}$, $K_{\text{lora}}$, and $K_{\text{context}}$, respectively; their order indicates the knowledge source at each hop.}
\label{fig:multihop-dual}
\end{figure*}

Figure~\ref{fig:multihop-dual} shows that the model selects the final entities of Chains~A and~B at similar rates across all knowledge-source assignments. If the model genuinely composed the three hops, a question about Chain~A should lead it to select the final entity of Chain~A substantially more often than that of Chain~B, with the opposite pattern for a question about Chain~B. But changing the queried chain does not consistently produce this corresponding change in the model's prediction. This indicates that the accuracy on \textsc{MhRBench} does not reflect successful composition of $K_{\text{base}}$, $K_{\text{context}}$, and $K_{\text{lora}}$. \textbf{Instead, the model appears to use the relation cues in the question to select an associated entity without composing the facts across the three hops.}

\subsection{Experiments on Two-hop Data}\label{main_results2}

To find out what causes the failure of composing three hops in Section~\ref{main_result3}, we next evaluate SHINE on the two-hop subset of \textsc{MhRBench}. The two-hop setting allows us to directly examine the pairwise composition ability among the three knowledge modules. Each instance combines two of $K_{\text{base}}$, $K_{\text{context}}$, and $K_{\text{lora}}$ and requires the model to pass the object of the first hop to the second hop to obtain the final answer.

\textbf{Table~\ref{tab:twohop-main} and Table~\ref{tab:twohop_dual} show that $K_{\text{base}}$ and $K_{\text{lora}}$ compose poorly in either order, but the reasons for failure are different.} When $K_{\text{base}}$ provides the first hop, the model predicts the intermediate entity in 95.00\% of instances but reaches the final answer in only 3.12\%. The dual-chain results show that this intermediate entity usually belongs to the queried chain, indicating that the model retrieves the correct parametric fact but fails to continue with the second-hop fact in $K_{\text{lora}}$. When $K_{\text{lora}}$ provides the first hop, the intermediate-entity rate rises to 97.00\%, while the final-answer rate falls to 0.25\%. In this case, however, the model selects the intermediate entities of the two chains at similar rates. Because the chains share the same relation, this pattern suggests that the adapter mainly matches the relation in the question and selects an associated entity without reliably using the starting entity to identify the relevant chain.

\begin{table*}[t]
  \centering
  \small
  \setlength{\tabcolsep}{9pt}
  \renewcommand{\arraystretch}{1.0}
  \setlength{\aboverulesep}{0.25ex}
  \setlength{\belowrulesep}{0.35ex}
  \setlength{\abovecaptionskip}{4pt}

  \begin{tabular}{ccccc}
  \toprule
  \multicolumn{2}{c}{Knowledge source assignment} &
  \multicolumn{3}{c}{Model prediction (\%)} \\
  \cmidrule(lr){1-2}\cmidrule(lr){3-5}
  Hop 1 source & Hop 2 source &
  Hop 1 entity & Answer & Unknown \\
  \midrule
  $K_{\text{base}}$ & $K_{\text{context}}$ & 5.75  & \textbf{80.12} & 14.13 \\
  $K_{\text{base}}$ & $K_{\text{lora}}$ & 95.00 & \textbf{3.12}  & 1.88  \\
  $K_{\text{context}}$ & $K_{\text{base}}$ & 0.00  & \textbf{83.00} & 17.00 \\
  $K_{\text{lora}}$ & $K_{\text{base}}$ & 97.00 & \textbf{0.25}  & 2.75  \\
  $K_{\text{lora}}$ & $K_{\text{context}}$ & 1.00  & \textbf{96.38} & 2.62 \\
  $K_{\text{context}}$ & $K_{\text{lora}}$ & 5.88  & \textbf{55.25} & 38.88 \\
  \bottomrule
  \end{tabular}

  \vspace{2mm}

  \parbox{0.94\textwidth}{
  \footnotesize
  \textbf{Note.}
  ``Hop 1'' denotes predictions matching the intermediate entity produced after the first hop, and ``Answer'' denotes predictions matching the entity reached after the second hop. ``Unknown'' denotes responses that match neither entity.
  }

  \caption{
  Results of SHINE on the two-hop subset of \textsc{MhRBench}.
  The first two columns show which knowledge source provides each hop.
  }
  \label{tab:twohop-main}
\end{table*}

\begin{table*}[t]
  \centering
  \small
  \setlength{\tabcolsep}{5.5pt}
  \renewcommand{\arraystretch}{1.0}
  \setlength{\aboverulesep}{0.25ex}
  \setlength{\belowrulesep}{0.35ex}
  \setlength{\abovecaptionskip}{4pt}
  \begin{tabular}{c c c c c c c c}
  \toprule
  \multicolumn{2}{c}{Knowledge source assignment} &
  & \multicolumn{5}{c}{Model prediction (\%)} \\
  \cmidrule(lr){1-2}\cmidrule(lr){4-8}
  Hop 1 & Hop 2 & Query &
  Hop1$_{\mathrm{A}}$ & Hop1$_{\mathrm{B}}$ &
  Answer$_{\mathrm{A}}$ & Answer$_{\mathrm{B}}$ &
  Unknown \\
  \midrule
  $K_{\text{base}}$    & $K_{\text{context}}$ & QA & 5.88  & 0.00  & \textbf{82.12} & 4.62  & 7.38 \\
  $K_{\text{base}}$    & $K_{\text{context}}$ & QB & 0.00  & 5.12  & 4.88  & \textbf{80.50} & 9.50 \\
  \addlinespace
  $K_{\text{base}}$    & $K_{\text{lora}}$    & QA & 89.25 & 4.00  & \textbf{1.25}  & 1.00  & 4.50 \\
  $K_{\text{base}}$    & $K_{\text{lora}}$    & QB & 7.12  & 86.12 & 1.12  & \textbf{0.75}  & 4.88 \\
  \addlinespace
  $K_{\text{context}}$ & $K_{\text{base}}$    & QA & 9.12  & 0.00  & \textbf{62.88} & 0.25  & 27.75 \\
  $K_{\text{context}}$ & $K_{\text{base}}$    & QB & 0.12  & 9.50  & 0.12  & \textbf{63.50} & 26.75 \\
  \addlinespace
  $K_{\text{lora}}$    & $K_{\text{base}}$    & QA & 47.62 & 46.12 & \textbf{1.50}  & 0.75  & 4.00 \\
  $K_{\text{lora}}$    & $K_{\text{base}}$    & QB & 45.00 & 49.00 & 1.25  & \textbf{0.88}  & 3.88 \\
  \addlinespace
  $K_{\text{lora}}$    & $K_{\text{context}}$ & QA & 3.88  & 5.00  & \textbf{1.50}  & 1.38  & 88.25 \\
  $K_{\text{lora}}$    & $K_{\text{context}}$ & QB & 4.00  & 4.50  & 1.62  & \textbf{1.88}  & 88.00 \\
  \addlinespace
  $K_{\text{context}}$ & $K_{\text{lora}}$    & QA & 13.12 & 0.00  & \textbf{14.50} & 3.75  & 68.62 \\
  $K_{\text{context}}$ & $K_{\text{lora}}$    & QB & 0.50  & 11.88 & 4.00  & \textbf{14.88} & 68.75 \\
  \bottomrule
  \end{tabular}
  \vspace{2mm}
  \parbox{0.94\textwidth}{
  \footnotesize
  \textbf{Note.}
  QA and QB are questions about chain A and chain B. Hop1$_{\mathrm{A}}$ and Hop1$_{\mathrm{B}}$ denote predictions matching the intermediate entity of each chain, and Answer$_{\mathrm{A}}$ and Answer$_{\mathrm{B}}$ denote predictions matching the entity at the second hop. ``Unknown'' matches neither chain.
  }
  \caption{
  Results of SHINE on the two-hop subset of \textsc{MhRBench-Dual}.
  }
  \label{tab:twohop_dual}
\end{table*}

\textbf{In contrast, base-model knowledge and textual context compose effectively in either order.} They achieve final-answer rates of 80.12\% and 83.00\%, respectively, and their dual-chain results show a clear preference for the final entity of the queried chain.

\textbf{Neither ordering shows reliable composition between $K_{\text{context}}$ and $K_{\text{lora}}$.}  With $K_{\text{context}}$ at the first hop and $K_{\text{lora}}$ at the second, final-answer accuracy is 55.25\% on \textsc{MhRBench}. On \textsc{MhRBench-Dual}, the model does not consistently select the queried chain: for QA, it chooses the answers from chains A and B in 14.50\% and 3.75\% of cases, respectively; for QB, the corresponding rates are 4.00\% and 14.88\%. This pattern suggests that the model has some ability to compose the two hops in this order, but this ability remains limited. In the reverse order, final-answer accuracy reaches 96.38\% on \textsc{MhRBench}, but falls to 1.50\% for QA and 1.88\% for QB on \textsc{MhRBench-Dual}. And the model produces the first-hop entities in most cases. The proportions of predictions matching the two chains' first-hop entities remain similar for both QA and QB.

The two-hop results show that $K_{\text{lora}}$ composes poorly with both $K_{\text{base}}$ and $K_{\text{context}}$, regardless of their order in the reasoning chain. This finding also explains the results in Section~\ref{main_result3}, which show that the model has poor ability to compose all three knowledge sources. Our longer-term goal is to find a hypernetwork-based continual learning system that composes knowledge well across sources.

\section{Conclusion}

In this work, we study how knowledge from the base model, input context, and hypernetwork-generated LoRA interacts under conflict and in multi-hop reasoning. On \textsc{TrCBench}, the model favors contextual knowledge, while stronger base-model priors increase the likelihood of retaining the original answer. We also identify source-specific hidden-state shifts that can steer the model’s source preference. On \textsc{MhRBench}, base-model knowledge and context compose well, but reasoning involving LoRA is less reliable. Dual-chain tests further show that some correct answers arise without following the links between facts. Together, these results reveal a gap between acquiring knowledge through an adapter and using it in reasoning across sources.

\newpage

\subsection*{AI use statement}

In this work, we used generative AI tools for generating conflicting facts in \textsc{TrCBench}, classifying model outputs, and assisting with translation. We have not used generative AI tools to help develop theoretical models or conceptual frameworks, design or provide feedback on research  methodology or experiments, implement methods, clean and reformat datasets, support qualitative and thematic data analysis or interpret results. The tasks of formulating mathematical claims, providing critical ingredients for proving mathematical claims, assisting in the writing of proofs and proposing or refining hypotheses are not applicable to this work. Additionally, we used generative AI tools to assist in coding, polish the research paper, search for information, summarize or analyze existing literature, identify relevant literature, and edit the draft to improve readability. We have reviewed all AI-assisted work. The LLM-generated conflict facts were manually verified. And the LLM-generated codes were also manually reviewed for correctness. We take responsibility for the final content of this work,
including text, claims or artifacts produced with the aid of generative AI.

\subsection*{Ethics statement}

This work does not involve human subjects, potentially harmful insights, fairness or legal issues, privacy concerns, or other ethical concerns. \textsc{TrCBench} contains synthetic conflict statements used solely to evaluate knowledge conflicts; they are not presented as factual claims. We filter and quality-check these statements and document how the benchmark is constructed.

\subsection*{Reproducibility statement}

We provide the inference code for all experiments and the complete data in the supplementary material. Appendix~\ref{app:trc_examples} to Appendix~\ref{app:dual_twohop_examples} also provides data examples and the corresponding experimental outputs for knowledge conflict, single-chain three-hop reasoning, dual-chain three-hop reasoning, single-chain two-hop reasoning, and dual-chain two-hop reasoning.

\newpage

\bibliography{iclr2027_conference}
\bibliographystyle{iclr2027_conference}

\newpage

\appendix

\section{Geometric Analysis of $K_{\text{context}}$ in Knowledge Conflict}\label{app:context_geometry}

\begin{figure}[H]
  \centering
  \includegraphics[width=0.80\linewidth]{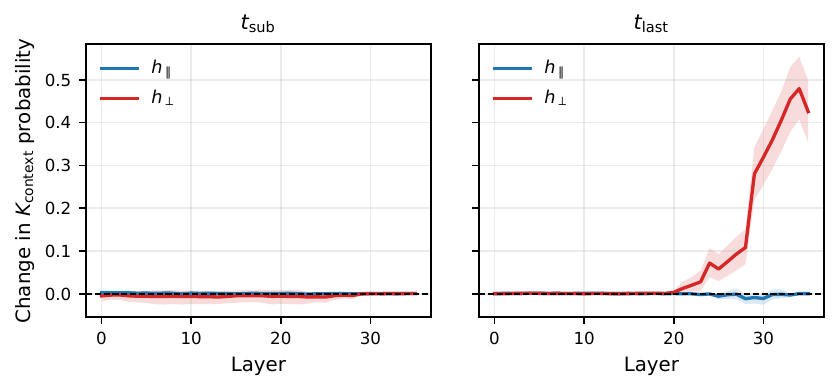}
  \caption{Causal effect of $h_{\parallel}$ and $h_{\perp}$ on the $K_{\text{context}}$ answer probability. The vertical axis is the change relative to the clean run. At each layer, the clean state at $t_{\mathrm{sub}}$ or $t_{\mathrm{last}}$ is replaced by $h_{\mathrm{clean}}+h_{\parallel}$ or $h_{\mathrm{clean}}+h_{\perp}$.}
  \label{fig:context-component-patch}
\end{figure}

\section{Analysis of Unknown Cases in the Three-hop Results of \textsc{MhRBench}}\label{app:unknown}

Each example is a three-hop chain $(s_0,r_1,o_1)$, $(o_1,r_2,o_2)$, $(o_2,r_3,o_3)$, and one knowledge source is assigned to each hop. \textcolor{entityblue}{\rule{0.7em}{0.7em}} is $s_0$, \textcolor{entitygreen}{\rule{0.7em}{0.7em}} is $o_1$, \textcolor{entityred}{\rule{0.7em}{0.7em}} is $o_2$, \textcolor{entityyellow}{\rule{0.7em}{0.7em}} is the answer $o_3$, and \textcolor{unknownpurple}{\rule{0.7em}{0.7em}} is an output that matches none of these entities.

The model answers with a name that is almost the hop-3 entity, but not the same string. \unknownoutput{Wamjol Myrfeol} differs from \chainfinal{Wamjov Myrfeol} by a single letter, and it is also distinct from \chainone{Microsoft} and the \chaintwo{Troarcae Compendium}. Because the output does not exactly match any hop entity, the case is counted as unknown.

\begin{multihopcase}{
    \(K_{\mathrm{base}}\rightarrow K_{\mathrm{context}}
    \rightarrow K_{\mathrm{lora}}\)}
  \textbf{$K_{\mathrm{base}}$:}
  \chainstart{Windows Server 2003} was developed by
  \chainone{Microsoft}.\par
  \textbf{$K_{\mathrm{lora}}$:}
  \chainfinal{Wamjov Myrfeol} is the appointed keeper of the
  \chaintwo{Troarcae Compendium}.\par
  \textbf{$K_{\mathrm{context}}$:}
  A reference to \chainone{Microsoft} appears in the
  \chaintwo{Troarcae Compendium}.\par
  \textbf{Question:}
  Who oversees the dossier containing a reference to the developer of
  \chainstart{Windows Server 2003}?\par
  \textbf{Output (Unknown):} \unknownoutput{Wamjol Myrfeol}\par
  \textbf{Gold answer:} \chainfinal{Wamjov Myrfeol}
\end{multihopcase}

The model combines pieces of two different hops. \unknownoutput{Skaern Vault} takes the name Skaern from the hop-2 record \chaintwo{Skaern Register} and the type word Vault from the hop-3 place \chainfinal{Seturn Vault}. The resulting string is not \chainone{Richard Adams}, not the register, and not the vault, so it is unknown.

\begin{multihopcase}{
    \(K_{\mathrm{base}}\rightarrow K_{\mathrm{context}}
    \rightarrow K_{\mathrm{lora}}\)}
  \textbf{$K_{\mathrm{base}}$:}
  The author of \chainstart{Watership Down} is
  \chainone{Richard Adams}.\par
  \textbf{$K_{\mathrm{lora}}$:}
  The \chaintwo{Skaern Register} is stored at
  \chainfinal{Seturn Vault}.\par
  \textbf{$K_{\mathrm{context}}$:}
  The \chaintwo{Skaern Register} was annotated by
  \chainone{Richard Adams}.\par
  \textbf{Question:}
  Where is the manuscript annotated by the author of
  \chainstart{Watership Down} stored?\par
  \textbf{Output (Unknown):} \unknownoutput{Skaern Vault}\par
  \textbf{Gold answer:} \chainfinal{Seturn Vault}
\end{multihopcase}

The model names an entity that occurs in the evidence but is not the object of any hop. The hop entities are \chainone{Microsoft}, the \chaintwo{Peskklav Collection}, and \chainfinal{Glaind Klaeth}. \unknownoutput{Kloarn Guild} matches none of them.

\begin{multihopcase}{
    \(K_{\mathrm{base}}\rightarrow K_{\mathrm{lora}}
    \rightarrow K_{\mathrm{context}}\)}
  \textbf{$K_{\mathrm{base}}$:}
  \chainstart{OLE DB} was developed by \chainone{Microsoft}.\par
  \textbf{$K_{\mathrm{lora}}$:}
  The \chaintwo{Peskklav Collection} mentions \chainone{Microsoft}.\par
  \textbf{$K_{\mathrm{context}}$:}
  Kloarn Guild appoints \chainfinal{Glaind Klaeth} to curate the
  \chaintwo{Peskklav Collection}.\par
  \textbf{Question:}
  Who oversees the dossier containing a reference to the developer of
  \chainstart{OLE DB}?\par
  \textbf{Output (Unknown):} \unknownoutput{Kloarn Guild}\par
  \textbf{Gold answer:} \chainfinal{Glaind Klaeth}
\end{multihopcase}

\section{Examples of \textsc{TrCBench} Results} \label{app:trc_examples}

Each example is one fact $(s,r,o)$. $K_{\text{base}}$, $K_{\text{context}}$, and $K_{\text{lora}}$ give three different objects for the same subject and relation, and the model answers the question. \textcolor{basegreen}{\rule{0.7em}{0.7em}} is the object from $K_{\text{base}}$, \textcolor{contextblue}{\rule{0.7em}{0.7em}} from $K_{\text{context}}$, \textcolor{lorared}{\rule{0.7em}{0.7em}} from $K_{\text{lora}}$, and \textcolor{unknownpurple}{\rule{0.7em}{0.7em}} is an output that matches none of them.

  \begin{trcexample}{Example 1: Answer $K_{\text{base}}$}
  \small
  \textbf{$K_{\text{base}}$:}
  IBM Parallel Sysplex is developed by \basehl{IBM}.\par
  \textbf{$K_{\text{context}}$:}
  IBM Parallel Sysplex is developed by \contexthl{Apple}.\par
  \textbf{$K_{\text{lora}}$:}
  IBM Parallel Sysplex is developed by \lorahl{Microsoft}.\par
  \textbf{Question:} IBM Parallel Sysplex is developed by\par
  \textbf{Output:} \basehl{IBM}
  \end{trcexample}

  \begin{trcexample}{Example 2: Answer $K_{\text{base}}$}
  \small
  \textbf{$K_{\text{base}}$:}
  Roman Republic is \basehl{Rome}.\par
  \textbf{$K_{\text{context}}$:}
  The capital of the Roman Republic is \contexthl{Carthage}.\par
  \textbf{$K_{\text{lora}}$:}
  The capital of the Roman Republic is \lorahl{Athens}.\par
  \textbf{Question:} What is the capital of Roman Republic?\par
  \textbf{Output:} \basehl{Rome}
  \end{trcexample}

  \begin{trcexample}{Example 3: Answer $K_{\text{lora}}$}
  \small
  \textbf{$K_{\text{base}}$:}
  Saudi Arabia is in \basehl{Asia}.\par
  \textbf{$K_{\text{context}}$:}
  Saudi Arabia is in \contexthl{Antarctica}.\par
  \textbf{$K_{\text{lora}}$:}
  Saudi Arabia is in \lorahl{Europe}.\par
  \textbf{Question:} Saudi Arabia is in\par
  \textbf{Output:} \lorahl{Europe}
  \end{trcexample}

  \begin{trcexample}{Example 4: Answer $K_{\text{lora}}$}
  \small
  \textbf{$K_{\text{base}}$:}
  Amos Anderson Art Museum is located in the country of
  \basehl{Finland}.\par
  \textbf{$K_{\text{context}}$:}
  Amos Anderson Art Museum is located in the country of
  \contexthl{Singapore}.\par
  \textbf{$K_{\text{lora}}$:}
  Amos Anderson Art Museum is located in the country of
  \lorahl{Brazil}.\par
  \textbf{Question:}
  Amos Anderson Art Museum is located in the country of\par
  \textbf{Output:} \lorahl{Brazil}
  \end{trcexample}

  \begin{trcexample}{Example 5: Answer $K_{\text{context}}$}
  \small
  \textbf{$K_{\text{base}}$:}
  The profession of Stafford Cripps is \basehl{politician}.\par
  \textbf{$K_{\text{context}}$:}
  The profession of Stafford Cripps is \contexthl{architect}.\par
  \textbf{$K_{\text{lora}}$:}
  The profession of Stafford Cripps is \lorahl{musician}.\par
  \textbf{Question:} The profession of Stafford Cripps is\par
  \textbf{Output:} \contexthl{Architect}
  \end{trcexample}

  \begin{trcexample}{Example 6: Answer $K_{\text{context}}$}
  \small
  \textbf{$K_{\text{base}}$:}
  Apocalypse Meow was created in the country of \basehl{Japan}.\par
  \textbf{$K_{\text{context}}$:}
  Apocalypse Meow was created in the country of \contexthl{France}.\par
  \textbf{$K_{\text{lora}}$:}
  Apocalypse Meow was created in the country of \lorahl{Italy}.\par
  \textbf{Question:}
  Apocalypse Meow was created in the country of\par
  \textbf{Output:} \contexthl{France}
  \end{trcexample}

  \begin{trcexample}{Example 7: Answer Unknown}
  \small
  \textbf{$K_{\text{base}}$:}
  What sport does Boris Diaw play? They play
  \basehl{basketball}.\par
  \textbf{$K_{\text{context}}$:}
  What sport does Boris Diaw play? They play
  \contexthl{football}.\par
  \textbf{$K_{\text{lora}}$:}
  What sport does Boris Diaw play? They play
  \lorahl{tennis}.\par
  \textbf{Question:} What sport does Boris Diaw play? They play\par
  \textbf{Output:} \unknownhl{Boris Diaw}
  \end{trcexample}

  \begin{trcexample}{Example 8: Answer Unknown}
  \small
  \textbf{$K_{\text{base}}$:}
  South Ossetia is \basehl{Tskhinvali}.\par
  \textbf{$K_{\text{context}}$:}
  The capital of South Ossetia is \contexthl{Moscow}.\par
  \textbf{$K_{\text{lora}}$:}
  South Ossetia \lorahl{does not have a capital city}.\par
  \textbf{Question:} What is the capital of South Ossetia?\par
  \textbf{Output:} \unknownhl{Tskarseti}
  \end{trcexample}

\section{Examples of Three-hop Results on \textsc{MhRBench}} \label{app:multihop_examples}

Each example is a three-hop chain $(s_0,r_1,o_1)$, $(o_1,r_2,o_2)$, $(o_2,r_3,o_3)$, with one knowledge source assigned to each hop. \textcolor{entityblue}{\rule{0.7em}{0.7em}} is $s_0$, \textcolor{entitygreen}{\rule{0.7em}{0.7em}} is $o_1$, \textcolor{entityred}{\rule{0.7em}{0.7em}} is $o_2$, \textcolor{entityyellow}{\rule{0.7em}{0.7em}} is the answer $o_3$, and \textcolor{unknownpurple}{\rule{0.7em}{0.7em}} matches none of these entities.

\begin{multihopcase}{Example 1:
    \(K_{\mathrm{base}}\rightarrow K_{\mathrm{lora}}
    \rightarrow K_{\mathrm{context}}\)}
  \textbf{$K_{\mathrm{base}}$:}
  \chainstart{Kinect} was developed by \chainone{Microsoft}.\par
  \textbf{$K_{\mathrm{lora}}$:}
  The \chaintwo{Shornveox Ledger} contains an entry about
  \chainone{Microsoft}.\par
  \textbf{$K_{\mathrm{context}}$:}
  Skairfraer Guild appoints \chainfinal{Spuev Braemspiv} to curate
  the \chaintwo{Shornveox Ledger}.\par
  \textbf{Question:}
  Who keeps the record that mentions the developer of
  \chainstart{Kinect}?\par
  \textbf{Output (Hop 1):} \chainone{Microsoft}\par
  \textbf{Gold answer:} \chainfinal{Spuev Braemspiv}
  \end{multihopcase}

  \begin{multihopcase}{Example 2:
    \(K_{\mathrm{base}}\rightarrow K_{\mathrm{context}}
    \rightarrow K_{\mathrm{lora}}\)}
  \textbf{$K_{\mathrm{base}}$:}
  The author of \chainstart{Medea} is \chainone{Euripides}.\par
  \textbf{$K_{\mathrm{lora}}$:}
  The \chaintwo{Quethvroask Codex} is stored at
  \chainfinal{Gyxkleond Vault}.\par
  \textbf{$K_{\mathrm{context}}$:}
  The \chaintwo{Quethvroask Codex} was annotated by
  \chainone{Euripides}.\par
  \textbf{Question:}
  Where is the manuscript annotated by the author of
  \chainstart{Medea} stored?\par
  \textbf{Output (Hop 3):} \chainfinal{Gyxkleond Vault}\par
  \textbf{Gold answer:} \chainfinal{Gyxkleond Vault}
  \end{multihopcase}

  \begin{multihopcase}{Example 3:
    \(K_{\mathrm{lora}}\rightarrow K_{\mathrm{base}}
    \rightarrow K_{\mathrm{context}}\)}
  \textbf{$K_{\mathrm{base}}$:}
  The official language of \chainone{Hungary} is
  \chaintwo{Hungarian}.\par
  \textbf{$K_{\mathrm{lora}}$:}
  \chainstart{Pryrncias Zhoalm} is a citizen of
  \chainone{Hungary}.\par
  \textbf{$K_{\mathrm{context}}$:}
  The designated manuscript for texts in \chaintwo{Hungarian}
  uses \chainfinal{Sheoxbri script}.\par
  \textbf{Question:}
  Which script does the designated manuscript for texts in the
  official language of \chainstart{Pryrncias Zhoalm}'s country of
  citizenship use?\par
  \textbf{Output (Hop 2):} \chaintwo{Hungarian}\par
  \textbf{Gold answer:} \chainfinal{Sheoxbri script}
  \end{multihopcase}

  \begin{multihopcase}{Example 4:
    \(K_{\mathrm{context}}\rightarrow K_{\mathrm{base}}
    \rightarrow K_{\mathrm{lora}}\)}
  \textbf{$K_{\mathrm{base}}$:}
  The capital of \chainone{Brazil} is \chaintwo{Brasília}.\par
  \textbf{$K_{\mathrm{lora}}$:}
  The designated hall in \chaintwo{Brasília} is built in the
  \chainfinal{Braevcliem school}.\par
  \textbf{$K_{\mathrm{context}}$:}
  \chainstart{Drund Jeskdaern} is a citizen of
  \chainone{Brazil}.\par
  \textbf{Question:}
  What is the architectural style of the designated hall in the
  capital of \chainstart{Drund Jeskdaern}'s country of citizenship?\par
  \textbf{Output (Hop 2):} \chaintwo{Brasília}\par
  \textbf{Gold answer:} \chainfinal{Braevcliem school}
  \end{multihopcase}

  \begin{multihopcase}{Example 5:
    \(K_{\mathrm{lora}}\rightarrow K_{\mathrm{context}}
    \rightarrow K_{\mathrm{base}}\)}
  \textbf{$K_{\mathrm{base}}$:}
  The capital of \chaintwo{Thailand} is
  \chainfinal{Bangkok}.\par
  \textbf{$K_{\mathrm{lora}}$:}
  \chainstart{Vrain Boxzhu} founded
  \chainone{Vundkien Guild}.\par
  \textbf{$K_{\mathrm{context}}$:}
  \chainone{Vundkien Guild} is located in
  \chaintwo{Thailand}.\par
  \textbf{Question:}
  What is the capital of the country where the guild founded by
  \chainstart{Vrain Boxzhu} is located?\par
  \textbf{Output (Hop 3):} \chainfinal{Bangkok}\par
  \textbf{Gold answer:} \chainfinal{Bangkok}
  \end{multihopcase}

  \begin{multihopcase}{Example6:
    \(K_{\mathrm{context}}\rightarrow K_{\mathrm{lora}}
    \rightarrow K_{\mathrm{base}}\)}
  \textbf{$K_{\mathrm{base}}$:}
  The capital of \chaintwo{Colombia} is
  \chainfinal{Bogotá}.\par
  \textbf{$K_{\mathrm{lora}}$:}
  \chainone{Voasktras Workshop} is located in
  \chaintwo{Colombia}.\par
  \textbf{$K_{\mathrm{context}}$:}
  \chainstart{Praern Skoax} founded
  \chainone{Voasktras Workshop}.\par
  \textbf{Question:}
  What is the capital of the country where the workshop founded by
  \chainstart{Praern Skoax} is located?\par
  \textbf{Output (Unknown):} \unknownoutput{Trondheim}\par
  \textbf{Gold answer:} \chainfinal{Bogotá}
  \end{multihopcase}

\section{Examples of Three-hop Results on \textsc{MhRBench-Dual}}
\label{app:dual_threehop_examples}

Each example has two chains, $(s_0,r_1,o_1)$, $(o_1,r_2,o_2)$, $(o_2,r_3,o_3)$ and $(s'_0,r_1,o'_1)$, $(o'_1,r_2,o'_2)$, $(o'_2,r_3,o'_3)$. In the first chain (Chain~A), \textcolor{achainblue}{\rule{0.7em}{0.7em}} is $s_0$, \textcolor{achaingreen}{\rule{0.7em}{0.7em}} is $o_1$, \textcolor{achainred}{\rule{0.7em}{0.7em}} is $o_2$, and \textcolor{achainyellow}{\rule{0.7em}{0.7em}} is the answer $o_3$. In the second chain (Chain~B): \textcolor{bchainblue}{\rule{0.7em}{0.7em}} is $s'_0$, \textcolor{bchaingreen}{\rule{0.7em}{0.7em}} is $o'_1$, \textcolor{bchainred}{\rule{0.7em}{0.7em}} is $o'_2$, and \textcolor{bchainyellow}{\rule{0.7em}{0.7em}} is the answer $o'_3$. \textcolor{unknownpurple}{\rule{0.7em}{0.7em}} matches none of them.

\begin{dualchaincase}{Example 1:
    \(K_{\mathrm{base}} \rightarrow K_{\mathrm{lora}} \rightarrow K_{\mathrm{context}}\)}
  \textbf{Chain A}\par
  \textbf{$K_{\mathrm{base}}$:}
  \Azero{Pinback} was created in the country of \Aone{United States of America}.\par
  \textbf{$K_{\mathrm{lora}}$:}
  The designated archive in \Aone{United States of America} is \Atwo{Cuthspuem Archive}.\par
  \textbf{$K_{\mathrm{context}}$:}
  The chief archivist of \Atwo{Cuthspuem Archive} is \Athree{Triandgail Frianbrie}.\par
  \textbf{Question A:}
  Who is the chief archivist of the designated archive in the country of origin of \Azero{Pinback}?\par
  \dualsep
  \textbf{Chain B}\par
  \textbf{$K_{\mathrm{base}}$:}
  \Bzero{Korean drama} was created in the country of \Bone{South Korea}.\par
  \textbf{$K_{\mathrm{lora}}$:}
  The designated archive in \Bone{South Korea} is \Btwo{Braxsur Archive}.\par
  \textbf{$K_{\mathrm{context}}$:}
  The chief archivist of \Btwo{Braxsur Archive} is \Bthree{Glaelm Shoalm}.\par
  \textbf{Question B:}
  Who is the chief archivist of the designated archive in the country of origin of \Bzero{Korean drama}?\par
  \dualsep
  \textbf{Answers}\par
  \textbf{QA Output (Unknown):} \unknownhl{Cuthspuem}\par
  \textbf{QA Answer:} \Athree{Triandgail Frianbrie}\par
  \textbf{QB Output (Hop~2, Chain~A):} \Atwo{Cuthspuem Archive}\par
  \textbf{QB Answer:} \Bthree{Glaelm Shoalm}\par
\end{dualchaincase}
\begin{dualchaincase}{Example 2:
    \(K_{\mathrm{base}} \rightarrow K_{\mathrm{context}} \rightarrow K_{\mathrm{lora}}\)}
  \textbf{Chain A}\par
  \textbf{$K_{\mathrm{base}}$:}
  The author of \Azero{Civil Disobedience} is \Aone{Henry David Thoreau}.\par
  \textbf{$K_{\mathrm{context}}$:}
  The \Atwo{Fraisroarn Collection} was annotated by \Aone{Henry David Thoreau}.\par
  \textbf{$K_{\mathrm{lora}}$:}
  The \Atwo{Fraisroarn Collection} is stored at \Athree{Priend Vault}.\par
  \textbf{Question A:}
  Where is the manuscript annotated by the author of \Azero{Civil Disobedience} stored?\par
  \dualsep
  \textbf{Chain B}\par
  \textbf{$K_{\mathrm{base}}$:}
  The author of \Bzero{Lady Chatterley's Lover} is \Bone{D. H. Lawrence}.\par
  \textbf{$K_{\mathrm{context}}$:}
  The \Btwo{Vrysglies Catalogue} was annotated by \Bone{D. H. Lawrence}.\par
  \textbf{$K_{\mathrm{lora}}$:}
  The \Btwo{Vrysglies Catalogue} is stored at \Bthree{Pableom Vault}.\par
  \textbf{Question B:}
  Where is the manuscript annotated by the author of \Bzero{Lady Chatterley's Lover} stored?\par
  \dualsep
  \textbf{Answers}\par
  \textbf{QA Output (Hop~2, Chain~A):} \Atwo{The Fraisroarn Collection}\par
  \textbf{QA Answer:} \Athree{Priend Vault}\par
  \textbf{QB Output (Hop~2, Chain~B):} \Btwo{The Vrysglies Catalogue}\par
  \textbf{QB Answer:} \Bthree{Pableom Vault}\par
\end{dualchaincase}
\begin{dualchaincase}{Example 3:
    \(K_{\mathrm{lora}} \rightarrow K_{\mathrm{base}} \rightarrow K_{\mathrm{context}}\)}
  \textbf{Chain A}\par
  \textbf{$K_{\mathrm{lora}}$:}
  \Azero{Nielm Biendvreon} is employed by \Aone{Mattel}.\par
  \textbf{$K_{\mathrm{base}}$:}
  The headquarters of \Aone{Mattel} is located in the city of \Atwo{El Segundo}.\par
  \textbf{$K_{\mathrm{context}}$:}
  The designated library in \Atwo{El Segundo} is built in the \Athree{Selmmer order}.\par
  \textbf{Question A:}
  What is the architectural style of the designated library in the headquarters city of \Azero{Nielm Biendvreon}'s employer?\par
  \dualsep
  \textbf{Chain B}\par
  \textbf{$K_{\mathrm{lora}}$:}
  \Bzero{Bloca Praevbarn} is employed by \Bone{Giant Eagle}.\par
  \textbf{$K_{\mathrm{base}}$:}
  The headquarters of \Bone{Giant Eagle} is located in the city of \Btwo{Pittsburgh}.\par
  \textbf{$K_{\mathrm{context}}$:}
  The designated library in \Btwo{Pittsburgh} is built in the \Bthree{Braindzhuem order}.\par
  \textbf{Question B:}
  What is the architectural style of the designated library in the headquarters city of \Bzero{Bloca Praevbarn}'s employer?\par
  \dualsep
  \textbf{Answers}\par
  \textbf{QA Output (Hop~1, Chain~B):} \Bone{Giant Eagle}\par
  \textbf{QA Answer:} \Athree{Selmmer order}\par
  \textbf{QB Output (Hop~1, Chain~A):} \Aone{Mattel}\par
  \textbf{QB Answer:} \Bthree{Braindzhuem order}\par
\end{dualchaincase}
\begin{dualchaincase}{Example 4:
    \(K_{\mathrm{context}} \rightarrow K_{\mathrm{base}} \rightarrow K_{\mathrm{lora}}\)}
  \textbf{Chain A}\par
  \textbf{$K_{\mathrm{context}}$:}
  \Azero{Shiemclien Telskaeth} is a citizen of \Aone{United States of America}.\par
  \textbf{$K_{\mathrm{base}}$:}
  The official language of \Aone{United States of America} is \Atwo{American English}.\par
  \textbf{$K_{\mathrm{lora}}$:}
  The designated manuscript for texts in \Atwo{American English} uses \Athree{Quyrfyx script}.\par
  \textbf{Question A:}
  Which script does the designated manuscript for texts in the official language of \Azero{Shiemclien Telskaeth}'s country of citizenship use?\par
  \dualsep
  \textbf{Chain B}\par
  \textbf{$K_{\mathrm{context}}$:}
  \Bzero{Hiand Voath} is a citizen of \Bone{Lieksa}.\par
  \textbf{$K_{\mathrm{base}}$:}
  The official language of \Bone{Lieksa} is \Btwo{Finnish}.\par
  \textbf{$K_{\mathrm{lora}}$:}
  The designated manuscript for texts in \Btwo{Finnish} uses \Bthree{Gluelgaelm script}.\par
  \textbf{Question B:}
  Which script does the designated manuscript for texts in the official language of \Bzero{Hiand Voath}'s country of citizenship use?\par
  \dualsep
  \textbf{Answers}\par
  \textbf{QA Output (Hop~3, Chain~A):} \Athree{Quyrfyx script}\par
  \textbf{QA Answer:} \Athree{Quyrfyx script}\par
  \textbf{QB Output (Hop~3, Chain~B):} \Bthree{Gluelgaelm script}\par
  \textbf{QB Answer:} \Bthree{Gluelgaelm script}\par
\end{dualchaincase}
\begin{dualchaincase}{Example 5:
    \(K_{\mathrm{lora}} \rightarrow K_{\mathrm{context}} \rightarrow K_{\mathrm{base}}\)}
  \textbf{Chain A}\par
  \textbf{$K_{\mathrm{lora}}$:}
  \Aone{Buncaen Conservancy} sponsors \Azero{Prain Prielmbron}.\par
  \textbf{$K_{\mathrm{context}}$:}
  The \Aone{Buncaen Conservancy} represents \Atwo{Ghana}.\par
  \textbf{$K_{\mathrm{base}}$:}
  \Atwo{Ghana} is located in the continent of \Athree{Africa}.\par
  \textbf{Question A:}
  What is the continent of the country represented by the conservancy sponsoring \Azero{Prain Prielmbron}?\par
  \dualsep
  \textbf{Chain B}\par
  \textbf{$K_{\mathrm{lora}}$:}
  \Bone{Thirnnaen Conservancy} sponsors \Bzero{Suern Trynkreolm}.\par
  \textbf{$K_{\mathrm{context}}$:}
  The \Bone{Thirnnaen Conservancy} represents \Btwo{Thailand}.\par
  \textbf{$K_{\mathrm{base}}$:}
  \Btwo{Thailand} is located in the continent of \Bthree{Asia}.\par
  \textbf{Question B:}
  What is the continent of the country represented by the conservancy sponsoring \Bzero{Suern Trynkreolm}?\par
  \dualsep
  \textbf{Answers}\par
  \textbf{QA Output (Hop~2, Chain~B):} \Btwo{Thailand}\par
  \textbf{QA Answer:} \Athree{Africa}\par
  \textbf{QB Output (Unknown):} \unknownhl{Europe}\par
  \textbf{QB Answer:} \Bthree{Asia}\par
\end{dualchaincase}
\begin{dualchaincase}{Example 6:
    \(K_{\mathrm{context}} \rightarrow K_{\mathrm{lora}} \rightarrow K_{\mathrm{base}}\)}
  \textbf{Chain A}\par
  \textbf{$K_{\mathrm{context}}$:}
  \Azero{Kloarspoan Kraim} founded \Aone{Nuklym Institute}.\par
  \textbf{$K_{\mathrm{lora}}$:}
  \Aone{Nuklym Institute} is located in \Atwo{Ireland}.\par
  \textbf{$K_{\mathrm{base}}$:}
  The official language of \Atwo{Ireland} is \Athree{Irish}.\par
  \textbf{Question A:}
  What is the official language of the country where the institute founded by \Azero{Kloarspoan Kraim} is located?\par
  \dualsep
  \textbf{Chain B}\par
  \textbf{$K_{\mathrm{context}}$:}
  \Bzero{Quurnkraesk Syxclyl} founded \Bone{Byxtrov Institute}.\par
  \textbf{$K_{\mathrm{lora}}$:}
  \Bone{Byxtrov Institute} is located in \Btwo{Kingdom of England}.\par
  \textbf{$K_{\mathrm{base}}$:}
  The official language of \Btwo{Kingdom of England} is \Bthree{English}.\par
  \textbf{Question B:}
  What is the official language of the country where the institute founded by \Bzero{Quurnkraesk Syxclyl} is located?\par
  \dualsep
  \textbf{Answers}\par
  \textbf{QA Output (Hop~3, Chain~B):} \Bthree{English}\par
  \textbf{QA Answer:} \Athree{Irish}\par
  \textbf{QB Output (Hop~1, Chain~B):} \Bone{Byxtrov Institute}\par
  \textbf{QB Answer:} \Bthree{English}\par
\end{dualchaincase}

\section{Examples of Two-Hop Results of \textsc{MhRBench}}
\label{app:twohop_examples}

Each example is a two-hop chain $(s_0,r_1,o_1)$, $(o_1,r_2,o_2)$, with one knowledge source assigned to each hop. \textcolor{entityblue}{\rule{0.7em}{0.7em}} is $s_0$, \textcolor{entitygreen}{\rule{0.7em}{0.7em}} is $o_1$, \textcolor{entityyellow}{\rule{0.7em}{0.7em}} is the answer $o_2$, and \textcolor{unknownpurple}{\rule{0.7em}{0.7em}} matches none of these entities.

\begin{multihopcase}{Example 1:
    \(K_{\mathrm{base}} \rightarrow K_{\mathrm{context}}\)}
  \textbf{$K_{\mathrm{base}}$:}
  The capital of \chainstart{Australia} is \chainone{Canberra}.\par
  \textbf{$K_{\mathrm{context}}$:}
  The \chainfinal{Musnyr Catalogue} has \chainone{Canberra} as its sole focus.\par
  \textbf{Question:}
  What is the name of the record whose sole focus is the capital of \chainstart{Australia}?\par
  \textbf{Output (Hop 2):} \chainfinal{The Musnyr Catalogue}\par
  \textbf{Gold answer:} \chainfinal{Musnyr Catalogue}
\end{multihopcase}
\begin{multihopcase}{Example 2:
    \(K_{\mathrm{base}} \rightarrow K_{\mathrm{lora}}\)}
  \textbf{$K_{\mathrm{base}}$:}
  The capital of \chainstart{Cuba} is \chainone{Havana}.\par
  \textbf{$K_{\mathrm{lora}}$:}
  The \chainfinal{Pruthglyth Dossier} has \chainone{Havana} as its sole focus.\par
  \textbf{Question:}
  Which record concerns only the capital of \chainstart{Cuba}?\par
  \textbf{Output (Hop 1):} \chainone{Havana}\par
  \textbf{Gold answer:} \chainfinal{Pruthglyth Dossier}
\end{multihopcase}
\begin{multihopcase}{Example 3:
    \(K_{\mathrm{lora}} \rightarrow K_{\mathrm{base}}\)}
  \textbf{$K_{\mathrm{base}}$:}
  The capital of \chainone{Morocco} is \chainfinal{Rabat}.\par
  \textbf{$K_{\mathrm{lora}}$:}
  The \chainstart{Climblond Register} has \chainone{Morocco} as its sole focus.\par
  \textbf{Question:}
  What is the capital of the entity that is the sole focus of \chainstart{Climblond Register}?\par
  \textbf{Output (Hop 1):} \chainone{Morocco}\par
  \textbf{Gold answer:} \chainfinal{Rabat}
\end{multihopcase}
\begin{multihopcase}{Example 4:
    \(K_{\mathrm{context}} \rightarrow K_{\mathrm{base}}\)}
  \textbf{$K_{\mathrm{base}}$:}
  \chainone{Guernica} was created by \chainfinal{Pablo Picasso}.\par
  \textbf{$K_{\mathrm{context}}$:}
  The \chainstart{Giexzhairn Compendium} has \chainone{Guernica} as its sole focus.\par
  \textbf{Question:}
  Who created the work that is the sole focus of \chainstart{Giexzhairn Compendium}?\par
  \textbf{Output (Unknown):} \unknownhl{Javier Marias}\par
  \textbf{Gold answer:} \chainfinal{Pablo Picasso}
\end{multihopcase}
\begin{multihopcase}{Example 5:
    \(K_{\mathrm{lora}} \rightarrow K_{\mathrm{context}}\)}
  \textbf{$K_{\mathrm{lora}}$:}
  The \chainstart{Ruethsel Corpus} is stored at \chainone{Wevkreosk Archive}.\par
  \textbf{$K_{\mathrm{context}}$:}
  The curator of \chainone{Wevkreosk Archive} is \chainfinal{Koask Spolklaen}.\par
  \textbf{Question:}
  Who is the curator of the archive where the \chainstart{Ruethsel Corpus} is stored?\par
  \textbf{Output (Hop 2):} \chainfinal{Koask Spolklaen}\par
  \textbf{Gold answer:} \chainfinal{Koask Spolklaen}
\end{multihopcase}
\begin{multihopcase}{Example 6:
    \(K_{\mathrm{context}} \rightarrow K_{\mathrm{lora}}\)}
  \textbf{$K_{\mathrm{lora}}$:}
  The director of \chainone{Fraer Archive} is \chainfinal{Praelm Zhilmbreov}.\par
  \textbf{$K_{\mathrm{context}}$:}
  The \chainstart{Wymzhi Folio} is stored at \chainone{Fraer Archive}.\par
  \textbf{Question:}
  Who is the director of the archive where the \chainstart{Wymzhi Folio} is stored?\par
  \textbf{Output (Unknown):} \unknownhl{Pilmbre}\par
  \textbf{Gold answer:} \chainfinal{Praelm Zhilmbreov}
\end{multihopcase}

\section{Examples of Two-Hop Results of \textsc{MhRBench-Dual}}
\label{app:dual_twohop_examples}

Each example has two chains, $(s_0,r_1,o_1)$, $(o_1,r_2,o_2)$ and $(s'_0,r_1,o'_1)$, $(o'_1,r_2,o'_2)$. In the first chain (Chain~A), \textcolor{achainblue}{\rule{0.7em}{0.7em}} is $s_0$, \textcolor{achaingreen}{\rule{0.7em}{0.7em}} is $o_1$, and \textcolor{achainyellow}{\rule{0.7em}{0.7em}} is the answer $o_2$. In the second chain (Chain~B), \textcolor{bchainblue}{\rule{0.7em}{0.7em}} is $s'_0$, \textcolor{bchaingreen}{\rule{0.7em}{0.7em}} is $o'_1$, and \textcolor{bchainyellow}{\rule{0.7em}{0.7em}} is the answer $o'_2$. \textcolor{unknownpurple}{\rule{0.7em}{0.7em}} matches none of them.

\begin{dualchaincase}{Example 1:
    \(K_{\mathrm{base}} \rightarrow K_{\mathrm{context}}\)}
  \textbf{Chain A}\par
  \textbf{$K_{\mathrm{base}}$:}
  The capital of \Azero{New Zealand} is \Aone{Wellington}.\par
  \textbf{$K_{\mathrm{context}}$:}
  The \Athree{Trairclon Folio} has \Aone{Wellington} as its sole focus.\par
  \textbf{Question A:}
  Identify the record focused solely on the capital of \Azero{New Zealand}.\par
  \dualsep
  \textbf{Chain B}\par
  \textbf{$K_{\mathrm{base}}$:}
  The capital of \Bzero{Portugal} is \Bone{Lisbon}.\par
  \textbf{$K_{\mathrm{context}}$:}
  The \Bthree{Zuequues Folio} has \Bone{Lisbon} as its sole focus.\par
  \textbf{Question B:}
  Identify the record focused solely on the capital of \Bzero{Portugal}.\par
  \dualsep
  \textbf{Answers}\par
  \textbf{QA Output (Hop~1, Chain~A):} \Aone{Wellington}\par
  \textbf{QA Answer:} \Athree{Trairclon Folio}\par
  \textbf{QB Output (Hop~2, Chain~B):} \Bthree{Zuequues Folio}\par
  \textbf{QB Answer:} \Bthree{Zuequues Folio}\par
\end{dualchaincase}
\begin{dualchaincase}{Example 2:
    \(K_{\mathrm{base}} \rightarrow K_{\mathrm{lora}}\)}
  \textbf{Chain A}\par
  \textbf{$K_{\mathrm{base}}$:}
  \Azero{Ludwig van Beethoven} was born in the city of \Aone{Bonn}.\par
  \textbf{$K_{\mathrm{lora}}$:}
  The \Athree{Zheoxdarn Reliquary} has \Aone{Bonn} as its sole focus.\par
  \textbf{Question A:}
  Identify the record focused solely on the birthplace of \Azero{Ludwig van Beethoven}.\par
  \dualsep
  \textbf{Chain B}\par
  \textbf{$K_{\mathrm{base}}$:}
  \Bzero{Charles Dickens} was born in the city of \Bone{Portsmouth}.\par
  \textbf{$K_{\mathrm{lora}}$:}
  The \Bthree{Felmkai Dossier} has \Bone{Portsmouth} as its sole focus.\par
  \textbf{Question B:}
  Identify the record focused solely on the birthplace of \Bzero{Charles Dickens}.\par
  \dualsep
  \textbf{Answers}\par
  \textbf{QA Output (Hop~1, Chain~A):} \Aone{Bonn}\par
  \textbf{QA Answer:} \Athree{Zheoxdarn Reliquary}\par
  \textbf{QB Output (Unknown):} \unknownhl{Bristol}\par
  \textbf{QB Answer:} \Bthree{Felmkai Dossier}\par
\end{dualchaincase}
\begin{dualchaincase}{Example 3:
    \(K_{\mathrm{lora}} \rightarrow K_{\mathrm{base}}\)}
  \textbf{Chain A}\par
  \textbf{$K_{\mathrm{lora}}$:}
  The \Azero{Zheoth Corpus} has \Aone{Johann Sebastian Bach} as its sole focus.\par
  \textbf{$K_{\mathrm{base}}$:}
  \Aone{Johann Sebastian Bach} was born in the city of \Athree{Eisenach}.\par
  \textbf{Question A:}
  Where was the person who is the sole focus of \Azero{Zheoth Corpus} born?\par
  \dualsep
  \textbf{Chain B}\par
  \textbf{$K_{\mathrm{lora}}$:}
  The \Bzero{Shaithvaeth Register} has \Bone{Ada Lovelace} as its sole focus.\par
  \textbf{$K_{\mathrm{base}}$:}
  \Bone{Ada Lovelace} was born in the city of \Bthree{London}.\par
  \textbf{Question B:}
  Where was the person who is the sole focus of \Bzero{Shaithvaeth Register} born?\par
  \dualsep
  \textbf{Answers}\par
  \textbf{QA Output (Hop~1, Chain~B):} \Bone{Ada Lovelace}\par
  \textbf{QA Answer:} \Athree{Eisenach}\par
  \textbf{QB Output (Hop~1, Chain~A):} \Aone{Johann Sebastian Bach}\par
  \textbf{QB Answer:} \Bthree{London}\par
\end{dualchaincase}
\begin{dualchaincase}{Example 4:
    \(K_{\mathrm{context}} \rightarrow K_{\mathrm{base}}\)}
  \textbf{Chain A}\par
  \textbf{$K_{\mathrm{context}}$:}
  The \Azero{Meosk Ledger} has \Aone{The Old Man and the Sea} as its sole focus.\par
  \textbf{$K_{\mathrm{base}}$:}
  The author of \Aone{The Old Man and the Sea} is \Athree{Ernest Hemingway}.\par
  \textbf{Question A:}
  Who is the author of the work that is the sole focus of \Azero{Meosk Ledger}?\par
  \dualsep
  \textbf{Chain B}\par
  \textbf{$K_{\mathrm{context}}$:}
  The \Bzero{Clalmlos Dossier} has \Bone{Hamlet} as its sole focus.\par
  \textbf{$K_{\mathrm{base}}$:}
  The author of \Bone{Hamlet} is \Bthree{William Shakespeare}.\par
  \textbf{Question B:}
  Who is the author of the work that is the sole focus of \Bzero{Clalmlos Dossier}?\par
  \dualsep
  \textbf{Answers}\par
  \textbf{QA Output (Hop~2, Chain~A):} \Athree{Ernest Hemingway}\par
  \textbf{QA Answer:} \Athree{Ernest Hemingway}\par
  \textbf{QB Output (Hop~2, Chain~B):} \Bthree{William Shakespeare}\par
  \textbf{QB Answer:} \Bthree{William Shakespeare}\par
\end{dualchaincase}
\begin{dualchaincase}{Example 5:
    \(K_{\mathrm{lora}} \rightarrow K_{\mathrm{context}}\)}
  \textbf{Chain A}\par
  \textbf{$K_{\mathrm{lora}}$:}
  The \Azero{Sorngliev Ledger} is stored at \Aone{Shasblier Archive}.\par
  \textbf{$K_{\mathrm{context}}$:}
  The chief archivist of \Aone{Shasblier Archive} is \Athree{Quurn Foathpain}.\par
  \textbf{Question A:}
  Who is the chief archivist of the archive where the \Azero{Sorngliev Ledger} is stored?\par
  \dualsep
  \textbf{Chain B}\par
  \textbf{$K_{\mathrm{lora}}$:}
  The \Bzero{Goathluesk Folio} is stored at \Bone{Noand Archive}.\par
  \textbf{$K_{\mathrm{context}}$:}
  The chief archivist of \Bone{Noand Archive} is \Bthree{Zuesk Cleonthyv}.\par
  \textbf{Question B:}
  Who is the chief archivist of the archive where the \Bzero{Goathluesk Folio} is stored?\par
  \dualsep
  \textbf{Answers}\par
  \textbf{QA Output (Hop~1, Chain~B):} \Bone{Noand Archive}\par
  \textbf{QA Answer:} \Athree{Quurn Foathpain}\par
  \textbf{QB Output (Hop~2, Chain~B):} \Bthree{Zuesk Cleonthyv}\par
  \textbf{QB Answer:} \Bthree{Zuesk Cleonthyv}\par
\end{dualchaincase}
\begin{dualchaincase}{Example 6:
    \(K_{\mathrm{context}} \rightarrow K_{\mathrm{lora}}\)}
  \textbf{Chain A}\par
  \textbf{$K_{\mathrm{context}}$:}
  The \Azero{Fiarprun Compendium} is stored at \Aone{Toaskeond Archive}.\par
  \textbf{$K_{\mathrm{lora}}$:}
  The superintendent of \Aone{Toaskeond Archive} is \Athree{Fraivlir Dryvven}.\par
  \textbf{Question A:}
  Who is the superintendent of the archive where the \Azero{Fiarprun Compendium} is stored?\par
  \dualsep
  \textbf{Chain B}\par
  \textbf{$K_{\mathrm{context}}$:}
  The \Bzero{Klyth Corpus} is stored at \Bone{Krairmaex Archive}.\par
  \textbf{$K_{\mathrm{lora}}$:}
  The superintendent of \Bone{Krairmaex Archive} is \Bthree{Zhoarfur Bumbrim}.\par
  \textbf{Question B:}
  Who is the superintendent of the archive where the \Bzero{Klyth Corpus} is stored?\par
  \dualsep
  \textbf{Answers}\par
  \textbf{QA Output (Hop~2, Chain~A):} \Athree{Fraivlir Dryvven}\par
  \textbf{QA Answer:} \Athree{Fraivlir Dryvven}\par
  \textbf{QB Output (Unknown):} \unknownhl{Frahoarfirum}\par
  \textbf{QB Answer:} \Bthree{Zhoarfur Bumbrim}\par
\end{dualchaincase}

\end{document}